\documentclass[a4paper,fleqn]{cas-dc}
 
\usepackage[numbers]{natbib}
\usepackage{algorithm}
\usepackage{algpseudocode}
\usepackage{listings} 
 
\def\tsc#1{\csdef{#1}{\textsc{\lowercase{#1}}\xspace}}
\tsc{WGM}
\tsc{QE}
\tsc{EP}
\tsc{PMS}
\tsc{BEC}
\tsc{DE}

\begin{document}
\let\WriteBookmarks\relax
\def\floatpagepagefraction{1}
\def\textpagefraction{.001}
\shorttitle{Evaluation of Sampling Strategies and Physics-Informed Kolmogorov--Arnold Networks in Unbounded Domains}
\shortauthors{Gregorio Pérez-Bernal et~al.}

\title [mode = title]{Evaluation of Sampling Strategies and Physics-Informed Kolmogorov--Arnold Networks in Unbounded Domains}                      
 
\author[1]{Gregorio Pérez-Bernal}[orcid=0009-0009-8586-7992] 

\author[1]{Oscar Rincón-Cardeño}[
                        orcid=0000-0002-5308-9710,
                        ]

\credit{Conceptualization of this study, Methodology, Software}

\affiliation[1]{organization={Mathematical Applications in Science and Engineering Research Group, School of Applied Sciences and Engineering, Universidad EAFIT},
                city={Medellín},
                country={Colombia}}

\author[2]{Silvana Montoya-Noguera}[%
   orcid={0000-0003-2490-9107}
   ]

\credit{Data curation, Writing - Original draft preparation}

\affiliation[2]{organization={Nature and Cities Research Group, School of Applied Sciences and Engineering, Universidad EAFIT},
                city={Medellín},
                country={Colombia}}

\author[1]{Nicolás Guarín-Zapata}[orcid=0000-0002-9435-1914]
 \cormark[1]
\ead{nguarinz@eafit.edu.co}

\cortext[cor1]{Corresponding author}

\begin{abstract}
Physics-informed neural networks (PINNs) have emerged as an effective approach for solving partial differential equations (PDEs) by incorporating physical laws into the learning process. However, their application to infinite and semi-infinite domains remains challenging due to the difficulty of representing unbounded regions with a finite number of training points. This work evaluates physics-informed learning strategies for inverse PDE problems in unbounded domains, focusing on the influence of sampling strategies based on uniform, Gaussian, and exponential distributions, and on the use of conventional multilayer perceptrons (MLPs) and Kolmogorov--Arnold Networks (KANs) within the PINN framework. The proposed benchmark considers manufactured inverse problems on infinite and semi-infinite domains, enabling quantitative assessment of reconstruction accuracy and computational efficiency. Results show that Gaussian and exponential sampling, controlled through explicit distribution parameters, improve reconstruction accuracy in the far field and offer a direct mechanism for embedding problem-specific prior knowledge into training. PIKANs, whose edge-based functional representation is grounded in the Kolmogorov--Arnold representation theorem, match or exceed the accuracy of MLP-based PINNs while requiring fewer neurons, though their spline-based formulation increases the computational cost of training, inference, and derivative evaluation.
\end{abstract}

\begin{graphicalabstract}
\includegraphics[width=6.7in]{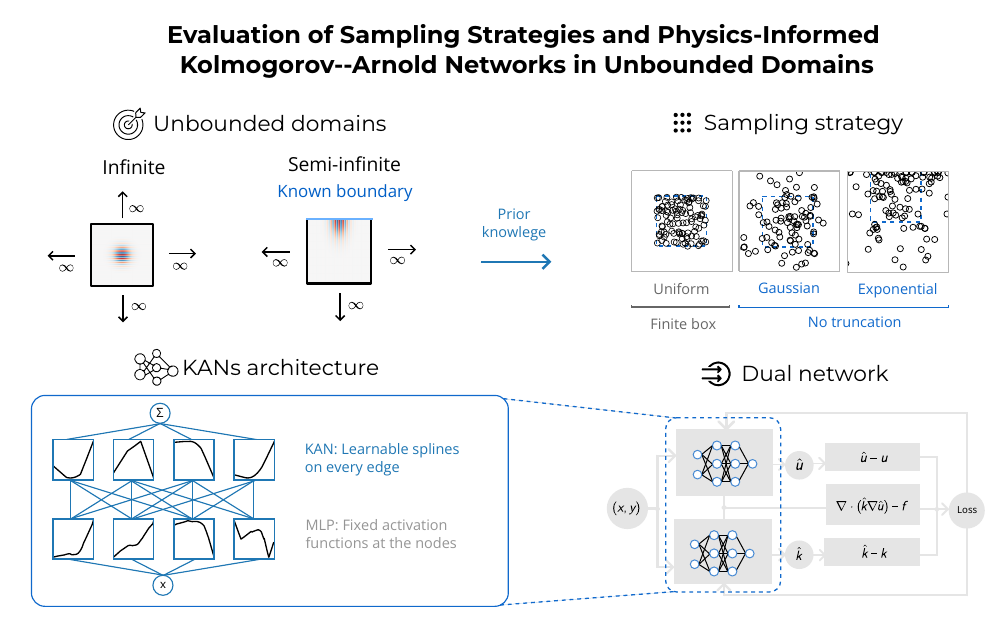}
\end{graphicalabstract}


\maketitle

\section{Introduction}\label{sec1}

Scientific machine learning (SciML) integrates scientific knowledge, such as governing equations, into machine learning models. Unlike purely data-driven approaches, SciML methods incorporate prior knowledge of the system to improve learning efficiency and reduce data requirements \citep{cuomoScientificMachineLearning2022}. This integration can be understood through three forms of bias: observational, inductive, and learning. Observational bias is introduced through the training data. Inductive bias arises from the model architecture and the assumptions imposed on the learning process. Learning bias is introduced through the training procedure, including the choice of loss function, optimization strategy, and sampling procedure \citep{rincon-cardeno_scoping_2026}. Among SciML approaches, physics-informed neural networks (PINNs) incorporate physical knowledge primarily through their loss formulation and training procedure. They have become a widely adopted framework for solving partial differential equations (PDEs) and inverse problems \cite{lytaev_numerical_2026}.

In the original formulation of PINNs, physical knowledge is incorporated into the loss function through the residual of the governing equations together with the corresponding boundary and, when applicable, initial conditions \citep{raissi2018physics}. This approach has shown promising results for forward and inverse problems, where unknown coefficients or physical parameters are inferred from observational data \citep{tarantolaInverse}. However, extending PINNs to infinite and semi-infinite domains remains challenging because an unbounded domain must be represented by a finite set of collocation points while preserving the asymptotic behavior of the solution. Numerical methods typically address this difficulty by truncating the computational domain and imposing artificial boundary conditions, although the resulting accuracy depends on the location of the truncation boundary and the imposed far-field conditions \citep{wangHighPrecisionPINNs2025}. For wave propagation problems, for example, the Sommerfeld radiation condition is commonly used to characterize the expected behavior of outgoing waves at infinity \citep{mesquita_numerical_2005,cardenoBenchmarkingPhysicsinformedNeural2026}.

Several aspects of the PINN formulation influence its performance on unbounded-domain problems. In particular, the sampling strategy determines the spatial distribution of collocation points and therefore the regions of the domain emphasized during training \citep{wu_comprehensive_2023}. Likewise, prior physical knowledge can be incorporated through regularization terms that encourage the expected asymptotic behavior of the solution without introducing artificial computational boundaries. The neural network architecture also plays an important role. For inverse problems, dual-network formulations employ separate neural networks to approximate the solution field and infer unknown parameters while being coupled through a common physics-informed objective function \citep{Zhang2023SeismicInversion}. Recently, Kolmogorov--Arnold Networks (KANs) have emerged as an alternative to conventional multilayer perceptrons (MLPs) for scientific machine learning due to their use of learnable spline-based activation functions \citep{liu2024kan}. While MLPs rely on the Universal Approximation Theorem \citep{hornik_multilayer_1989}, KANs are motivated by the Kolmogorov--Arnold representation theorem, which states that multivariate continuous functions can be represented as finite compositions of univariate functions \citep{schmidt-hieber_kolmogorovarnold_2021}. Their different approximation mechanisms have motivated recent studies comparing both architectures across a variety of scientific computing applications. The combined effects of sampling strategies and network architectures, such as KANs, on the accuracy and computational cost of PINNs in unbounded domains have not been evaluated. In this work, we address this research gap by systematically evaluating these complementary strategies within a dual-network framework, considering both KANs and MLPs for inverse problems in infinite and semi-infinite domains. Specifically, we address the following research question:

\textit{How can we leverage sampling strategies and network architectures to improve the performance of physics-informed neural networks in unbounded domain problems?}\\

To answer this, we present an evaluation of physics-informed learning strategies for inverse problems posed on infinite and semi-infinite domains. Specifically, we investigate the influence of domain sampling distributions, such as Gaussian and exponential, to promote the expected far-field behavior, and compare conventional MLP-based PINNs with Physics-Informed Kolmogorov--Arnold Networks (PIKANs). Two manufactured inverse problems serve as benchmarks, enabling a quantitative assessment of parameter inference accuracy and computational efficiency to guide effective model design. The findings of this study can provide guidance on the selection of the sampling distribution and the network architecture for inverse problems in unbounded domains. The remainder of this paper is organized as follows. Section~\ref{sec:methodology} presents the dual-network framework, the Gaussian and exponential sampling strategies, and the MLP and KAN architectures. Section~\ref{sec:problem} defines the manufactured inverse problems on the infinite and semi-infinite domains. Section~\ref{sec:hyperparameters} describes the hyperparameter optimization of the networks and the adaptive weighting of the loss terms. Section~\ref{sec:sampling} evaluates the effect of the sampling strategies on the reconstruction of the solution and the coefficient inside and outside the training region. Section~\ref{sec:pikans} analyzes the trade-off between accuracy, model size, and computational cost for MLPs and PIKANs.

\section{Strategies for physics-informed learning in unbounded domains}\label{sec:methodology}

\begin{figure*}[ht]
    \centering
    \includegraphics[width=6.3 in]{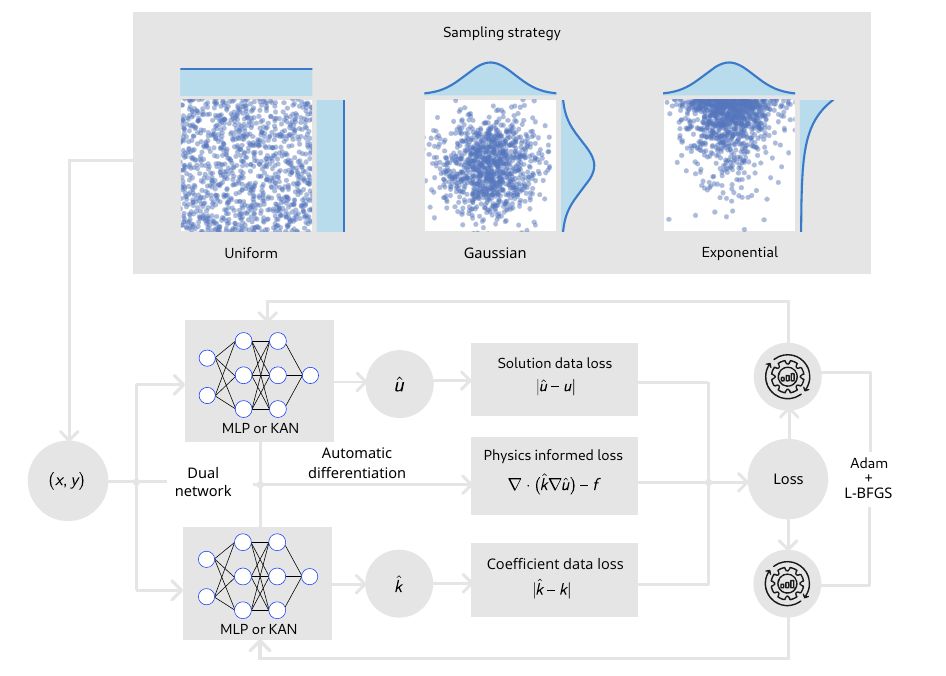}
\caption{Schematic of the proposed dual-network physics-informed framework.  The coordinates $(x,y)$ are provided to two neural networks that simultaneously approximate the solution $\hat{u}$ and the unknown coefficient $\hat{k}$.  We used uniform, Gaussian, and exponential sampling to distribute the PDE collocation points according to the computational domain. Automatic differentiation is used to evaluate the PDE residual, while solution and coefficient data losses are combined with the PDE 
residual to construct the composite loss. The two networks are jointly optimized  by minimizing the resulting objective using Adam followed by L-BFGS.}
    \label{fig:dual_network}
\end{figure*}

In a standard PINN formulation, a neural network is used to approximate the solution of a differential equation as
\begin{equation*}
    \mathcal{NN}(\mathbf{x};\theta) \approx u(\mathbf{x}),
\end{equation*}
where $u(\mathbf{x})$ denotes the solution of the governing differential equation, $\mathbf{x}\in\Omega$ represents the spatial coordinates, and $\theta$ denotes the trainable parameters of the network. Applying physics-informed learning to inverse problems posed on unbounded domains introduces additional challenges. Since the computational domain cannot be sampled uniformly over an infinite region, the choice of collocation points becomes relevant for accurately representing both the region where the solution exhibits significant variations and its asymptotic behavior. Moreover, inverse problems require the simultaneous approximation of the solution field and the identification of unknown spatially varying parameters, resulting in a more demanding optimization problem. The workflow of the proposed framework is illustrated in Fig.~\ref{fig:dual_network}. This work evaluates complementary strategies to the standard PINN formulation for addressing these challenges. Specifically, we consider (i) a dual-network architecture for the simultaneous approximation of the solution and the unknown coefficients map, (ii) alternative sampling based on uniform, Gaussian, and exponential distributions to improve the representation of unbounded domains, and (iii) KANs as the underlying neural network architecture. The training workflow adopted in this study is summarized in Algorithm~\ref{alg:proposed}.

\begin{algorithm}[ht]
\caption{\small\sffamily Training of physics-informed learning strategies in unbounded domains}
\label{alg:proposed}

\small\sffamily

\textbf{Inputs:}\\
Governing PDE on an infinite or semi-infinite domain\\
Set of observation and boundary data\\
Sampling strategy
$\in\{\mathrm{Uniform},\mathrm{Gaussian},\mathrm{Exponential}\}$\\
Neural network architecture
$\in\{\mathrm{MLP},\mathrm{KAN}\}$

\vspace{2mm}

\textbf{Outputs:}\\
Predicted solution $\hat{u}(\mathbf{x};\theta_u)$\\
Predicted coefficient $\hat{k}(\mathbf{x};\theta_k)$

\begin{algorithmic}

\State Initialize $\theta_u$, $\theta_k$, loss weights,
optimizers, and stopping criteria.

\State Generate PDE collocation points according to the selected
sampling strategy.

\State Generate observation points and, for the semi-infinite problem,
boundary points.

\For{optimizer $\in \{\mathrm{Adam},\mathrm{L\mbox{-}BFGS}\}$}

    \While{stopping criterion is not satisfied}

        \State Evaluate $\hat{u}$ and $\hat{k}$.

        \State Compute the PDE residual using automatic differentiation.

        \State Compute the data and PDE loss terms.

        \State Assemble the corresponding composite loss.

        \State Update the loss weights adaptively.

        \State Update $\theta_u$ and $\theta_k$.

    \EndWhile

\EndFor

\State \Return $\hat{u}$ and $\hat{k}$.

\end{algorithmic}
\end{algorithm}

\textbf{Dual-network architecture:} Instead of a single network with multiple outputs, we employ two independent neural networks following \cite{Zhang2023SeismicInversion},
\begin{equation*}
    \mathcal{NN}^{u}(\mathbf{x};\theta_u) \approx u(\mathbf{x}),
    \qquad
    \mathcal{NN}^{k}(\mathbf{x};\theta_k) \approx k(\mathbf{x}),
\end{equation*}
where $\theta_u$ and $\theta_k$ are the trainable parameters. This dual-network
formulation allows each field to have its own trainable representation while
preserving their physical coupling through the governing equation. In contrast
to a single multi-output network, the representation of each field is learned
independently, which is useful here because $u$ and $k$ have different roles in
the inverse problem and may exhibit different spatial behavior. The two
networks are therefore not optimized independently: their predictions are
combined in the physics-informed loss, and the PDE residual couples the
reconstruction of $u$ and $k$. Automatic differentiation is used to compute the
spatial derivatives required by the governing-equation residual
\citep{hare2023autograd}.

The two networks are trained using PDE collocation points
$\{\mathbf{x}_i^{\mathrm{PDE}}\}_{i=1}^{N_{\mathrm{PDE}}}$,
observation points
$\{\mathbf{x}_i^{\mathrm{obs}}\}_{i=1}^{N_{\mathrm{obs}}}$, and, for the semi-infinite problem, boundary points
$\{\mathbf{x}_i^{\mathrm{bnd}}\}_{i=1}^{N_{\mathrm{bnd}}}$.
We use $N_{\mathrm{obs}}=100$ observation points to represent an inverse problem with limited data.
The PDE is enforced at $N_{\mathrm{PDE}}=10{,}000$ collocation points, so the reconstruction is constrained by the governing equation over the domain rather than by observations alone.
For the semi-infinite problem, we use $N_{\mathrm{bnd}}=10$ boundary points.
The PDE residual loss is defined as
\begin{equation*}
\mathcal{L}_{\mathrm{PDE}}
=
\frac{1}{N_{\mathrm{PDE}}}
\sum_{i=1}^{N_{\mathrm{PDE}}}
\left|
\mathcal{R}\!\left(\mathbf{x}_i^{\mathrm{PDE}}\right)
\right|^2,
\end{equation*}
where $\mathcal{R}$ denotes the governing-equation residual. The observation losses for the state and unknown coefficients are
\begin{equation*}
\mathcal{L}_{u}
=
\frac{1}{N_{\mathrm{obs}}}
\sum_{i=1}^{N_{\mathrm{obs}}}
\left|
\hat{u}\!\left(\mathbf{x}_i^{\mathrm{obs}}\right)
-
u\!\left(\mathbf{x}_i^{\mathrm{obs}}\right)
\right|^2,
\end{equation*}
and
\begin{equation*}
\mathcal{L}_{k}
=
\frac{1}{N_{\mathrm{obs}}}
\sum_{i=1}^{N_{\mathrm{obs}}}
\left|
\hat{k}\!\left(\mathbf{x}_i^{\mathrm{obs}}\right)
-
k\!\left(\mathbf{x}_i^{\mathrm{obs}}\right)
\right|^2.
\end{equation*}
For the semi-infinite domain, the Dirichlet boundary condition is additionally enforced through
\begin{equation*}
\mathcal{L}_{\mathrm{bnd}}
=
\frac{1}{N_{\mathrm{bnd}}}
\sum_{i=1}^{N_{\mathrm{bnd}}}
\left|
\hat{u}\!\left(\mathbf{x}_i^{\mathrm{bnd}}\right)
-
u\!\left(\mathbf{x}_i^{\mathrm{bnd}}\right)
\right|^2.
\end{equation*}

The network parameters are obtained by minimizing the corresponding composite loss,
\begin{equation*}
(\theta_u^{*},\theta_k^{*})
=
\arg\min_{\theta_u,\theta_k}
\mathcal{L}(\theta_u,\theta_k),
\end{equation*}
where
\begin{equation*}
\mathcal{L}_{\mathrm{inf}}
=
\lambda_{\mathrm{PDE}}\mathcal{L}_{\mathrm{PDE}}
+
\lambda_{u}\mathcal{L}_{u}
+
\lambda_{k}\mathcal{L}_{k},
\end{equation*}
for the infinite domain, and
\begin{equation*}
\mathcal{L}_{\mathrm{semi}}
=
\lambda_{\mathrm{PDE}}\mathcal{L}_{\mathrm{PDE}}
+
\lambda_{u}\mathcal{L}_{u}
+
\lambda_{k}\mathcal{L}_{k}
+
\lambda_{\mathrm{bnd}}\mathcal{L}_{\mathrm{bnd}},
\end{equation*}
for the semi-infinite domain. The boundary points used to evaluate
$\mathcal{L}_{\mathrm{bnd}}$ are sampled from a Gaussian distribution,
consistent with the proposed sampling strategy. The coefficients
$\lambda_{\mathrm{PDE}}$, $\lambda_u$, $\lambda_k$, and $\lambda_{\mathrm{bnd}}$
control the relative contribution of each term. Following the original PINNs
formulation, they are initialized to one \citep{raissi2018physics} and
subsequently adjusted using the adaptive weighting strategy of
\cite{gao_physics-informed_2025}. The optimization of loss weights with scheduled and weighting strategies is discussed in Supplementary Material (Section~S2, Adaptive weighting).\\
 
\textbf{Gaussian and exponential sampling strategy:}
Sampling strategies have become an important approach for improving the
convergence and accuracy of physics-informed neural networks
\citep{luo_physics-informed_2025}. While standard PINNs typically sample
training points uniformly within a prescribed finite region, our approach
defines their spatial distribution through a prescribed probability density
$\rho(\mathbf{x})$. This provides a way to incorporate prior knowledge about
the spatial behavior of the solution into the training process without
defining a finite truncation region. Assuming a localized solution that
decays toward infinity,
\begin{equation*}
    \lim_{\|\mathbf{x}\|\rightarrow\infty}u(\mathbf{x})=0,
\end{equation*}
the sampling density can assign more training points to regions where
larger spatial variations are expected and fewer points to regions where
the solution approaches its asymptotic state. The parameters of the
distributions control the spatial extent of the sampled region and thus
determine how training points are distributed over the domain. For the infinite-domain problem, independent Gaussian distributions are
used along both spatial directions,
\begin{equation*}
    \rho_G(x,y)=\rho_x(x)\rho_y(y),
\end{equation*}
where
\begin{equation*}
\begin{aligned}
\rho_x(x)
&=\frac{1}{\sqrt{2\pi}\sigma_x}
\exp\left[-\frac{(x-\mu_x)^2}{2\sigma_x^2}\right],\\
\rho_y(y)
&=\frac{1}{\sqrt{2\pi}\sigma_y}
\exp\left[-\frac{(y-\mu_y)^2}{2\sigma_y^2}\right].
\end{aligned}
\end{equation*}
Here, $\mu_x$ and $\mu_y$ define the center of the sampling distribution,
while $\sigma_x$ and $\sigma_y$ control its spread along each direction.
The training points are sampled according to
\begin{equation*}
    \mathbf{x}_i=(x_i,y_i)\sim\rho_G,
    \qquad i=1,\ldots,N,
\end{equation*}
where $N$ denotes the number of training points generated from the
distribution. This sampling concentrates points around the region of
interest while retaining a non-zero probability of sampling points farther
from the origin. In contrast to uniform sampling over a predefined
finite region, the Gaussian distribution provides a continuous sampling
density over the unbounded domain, with the spatial concentration
controlled by $\sigma_x$ and $\sigma_y$. For the semi-infinite problem, the sampling density combines a Gaussian
distribution along the unbounded direction with an exponential
distribution along the semi-infinite direction,
\begin{equation*}
    \rho_H(x,y)=\rho_x(x)\rho_E(y),
\end{equation*}
where, for the physical domain $y\leq0$,
\begin{equation*}
    \rho_E(y)=\lambda e^{\lambda y},
    \qquad y\leq0.
\end{equation*}
Here, $\sigma_x$ controls the spread of the samples along the unbounded
direction, while $\lambda$ controls their decay along the semi-infinite
direction. The training points are sampled according to
\begin{equation*}
    \mathbf{x}_i=(x_i,y_i)\sim\rho_H,
    \qquad i=1,\ldots,N.
\end{equation*}
The resulting distribution places more training points near the boundary
and progressively fewer points toward the asymptotic region. Therefore,
$\sigma_x$ and $\lambda$ provide direct control of the spatial distribution
of the training points without requiring a predefined finite region for
the unbounded directions such as in the case of uniform sampling.\\

\begin{figure*}[ht!]
    \centering
    \includegraphics[width=6.3 in]{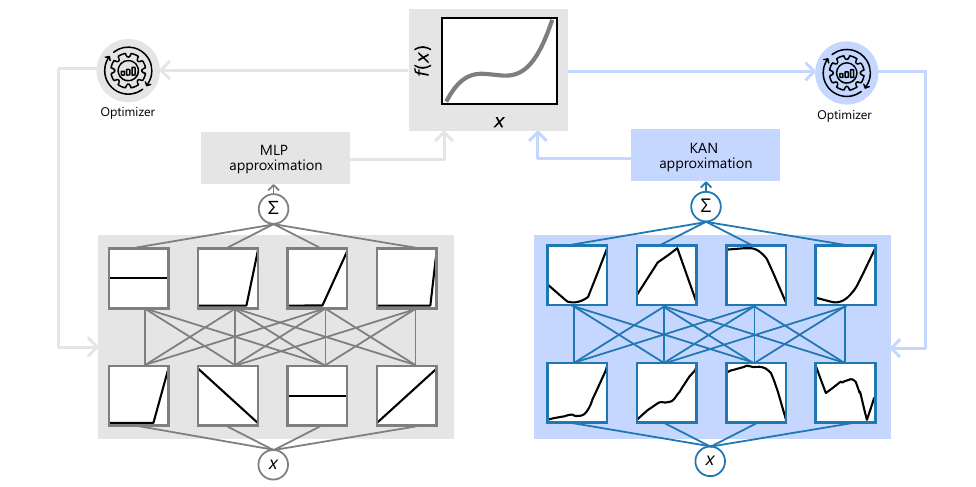}
    \caption{Conceptual comparison between multilayer perceptrons (MLPs) and Kolmogorov--Arnold networks (KANs) for function approximation. In the MLP schematic (left), nonlinear transformations are illustrated at the nodes using a fixed activation function. ReLU is used only as an illustrative example in the visualization. In contrast, a KAN (right) places learnable nonlinear functions on the network edges, allowing each connection to adapt its transformation.}
    \label{fig:MLP_KAN}
\end{figure*}

\textbf{MLP and KAN architectures:} To evaluate the neural network architecture influence in unbounded domains, the proposed framework is implemented using two different architectures: standard MLPs and KANs. As illustrated in Fig.~\ref{fig:MLP_KAN}, the two architectures differ fundamentally in how they represent nonlinear mappings. The panels depict representative learned edge functions, highlighting the architectural difference between both models while approximating the same target function.

The MLP serves as the reference architecture because it constitutes a standard
architecture used in PINN formulations. An MLP approximates the solution
through a sequence of linear transformations followed by fixed nonlinear
activation functions. Its expressive capacity is primarily controlled by the
network depth, width, and choice of activation function. Consequently, the
principal architectural hyperparameters correspond to the number of hidden
layers, the number of neurons per layer, and the activation function. In
contrast, KANs replace fixed node activations with learnable spline-based
functions defined on the network edges according to the Kolmogorov--Arnold
representation theorem \citep{liu2024kan}. This formulation introduces
additional architecture-specific hyperparameters, particularly the spline
grid size and spline order, which control the resolution and smoothness of
the learned edge functions. These parameters influence both the approximation
capability and the computational complexity of the network.

By replacing fixed activation functions with learnable spline-based edge
functions, KANs seek to exploit this representation to provide greater
flexibility in approximating complex nonlinear mappings. Since the solutions
considered in this work combine localized oscillatory behavior with smooth
asymptotic decay, a more expressive functional representation may better
capture these distinct regimes without requiring excessively large networks.
Therefore, our working hypothesis is that the adaptive spline representation
employed by KANs can provide a more flexible approximation of both the solution
field and the unknown coefficient than conventional MLPs. At the same time,
the additional computational cost associated with optimizing spline functions
must be assessed for a complete analysis. To quantify this cost, we compare the
forward operations and the trainable parameters of both architectures. For an MLP layer with $n_{\mathrm{in}}$ input units and $n_{\mathrm{out}}$
output units, the output of neuron $o$ is
\begin{equation*}
z_o^{l}
=
\phi\left(
\sum_{i=1}^{n_{\mathrm{in}}}
w_{o,i}^{l}x_i^{l-1}
+
b_o^{l}
\right),
\qquad
o=1,\ldots,n_{\mathrm{out}},
\end{equation*}
where $x_i^{l-1}$ is the $i$-th input, $w_{o,i}^{l}$ is the corresponding
weight, $b_o^{l}$ is the bias of the $o$-th output unit, and $\phi$ is the
activation function. In a KAN layer with the same input and output dimensions,
the scalar weights and the fixed activation are replaced by learnable
univariate functions on the edges. We use a PyTorch implementation
of KANs \citep{blealtan2024efficientkan}, in which the output of neuron $o$ is
\begin{equation*}
z_o^{l}
=
\sum_{i=1}^{n_{\mathrm{in}}}
\left[
w_{o,i}^{l}\phi\left(x_i^{l-1}\right)
+
s_{o,i}^{l}
\sum_{g=1}^{G+k}
c_{o,i,g}^{l}
B_g\left(x_i^{l-1}\right)
\right],
\end{equation*}
for $o=1,\ldots,n_{\mathrm{out}}$. Here, $\phi$ is the fixed base activation, $B_g$ is the $g$-th B-spline basis
function, $G$ is the grid size, and $k$ is the spline order. The trainable
quantities are the base weights $w^l$, the spline coefficients $c^l$, and the
spline scaling factors $s^l$. We consider networks with input dimension $d_{\mathrm{in}}$, output dimension
$d_{\mathrm{out}}$, and $L$ hidden layers of width $N$. For an MLP, the number
of trainable parameters is
\begin{equation*}
P_{\mathrm{MLP}}
=
(d_{\mathrm{in}}+1)N
+
(L-1)\,N(N+1)
+
(N+1)d_{\mathrm{out}}.
\end{equation*}
The forward pass consists of matrix multiplications with
$d_{\mathrm{in}}N$ operations in the first layer, $N^2$ operations in each of
the $L-1$ hidden-to-hidden layers, and $Nd_{\mathrm{out}}$ operations in the
output layer. Using the convention of two FLOPs per multiply--accumulate
operation, the FLOP count is
\begin{equation*}
F_{\mathrm{MLP}}
=
2\left[
d_{\mathrm{in}}N
+
(L-1)N^2
+
Nd_{\mathrm{out}}
\right],
\end{equation*}
which grows quadratically with the layer width. For a KAN layer connecting
$n_{l-1}$ input units to $n_l$ output units, each edge contains one base
weight, one spline scaling factor, and $G+k$ spline coefficients. The number of
trainable parameters is therefore
\begin{equation*}
P_{\mathrm{KAN}}
=
(G+k+2)
\sum_{l=1}^{L+1}
n_{l-1}n_l,
\end{equation*}
with $n_0=d_{\mathrm{in}}$, $n_{L+1}=d_{\mathrm{out}}$, and $n_l=N$ for the
hidden layers. The forward computation involves two matrix multiplications:
one for the base weights, with $n_{l-1}n_l$ multiply--accumulate operations,
and one for the spline coefficients, with $n_{l-1}(G+k)n_l$ operations, since
each input is represented by $G+k$ basis values. The total FLOP count is
\begin{equation*}
F_{\mathrm{KAN}}
=
2(G+k+1)
\sum_{l=1}^{L+1}
n_{l-1}n_l.
\end{equation*}
Compared with an MLP layer of the same dimensions, a KAN layer requires
$G+k+1$ times more FLOPs per edge and $G+k+2$ times more parameters per edge.
Both counts scale with the product $n_{l-1}n_l$, so a KAN with fewer neurons
per layer can have a parameter count comparable to that of an MLP. These counts
include only the matrix multiplications and do not include the evaluation of
the B-spline basis functions. Detailed profiling can be performed with tools
such as \texttt{FlopCountAnalysis} from \texttt{fvcore} \citep{fvcore}.
Consequently, the experimental evaluation compares both architectures in terms
of solution accuracy, number of trainable parameters, and computational cost.\\

\section{Manufactured-solution-based problems in unbounded domains}\label{sec:problem}

\begin{figure*}[ht]
    \centering
    \includegraphics[width=6.3 in]{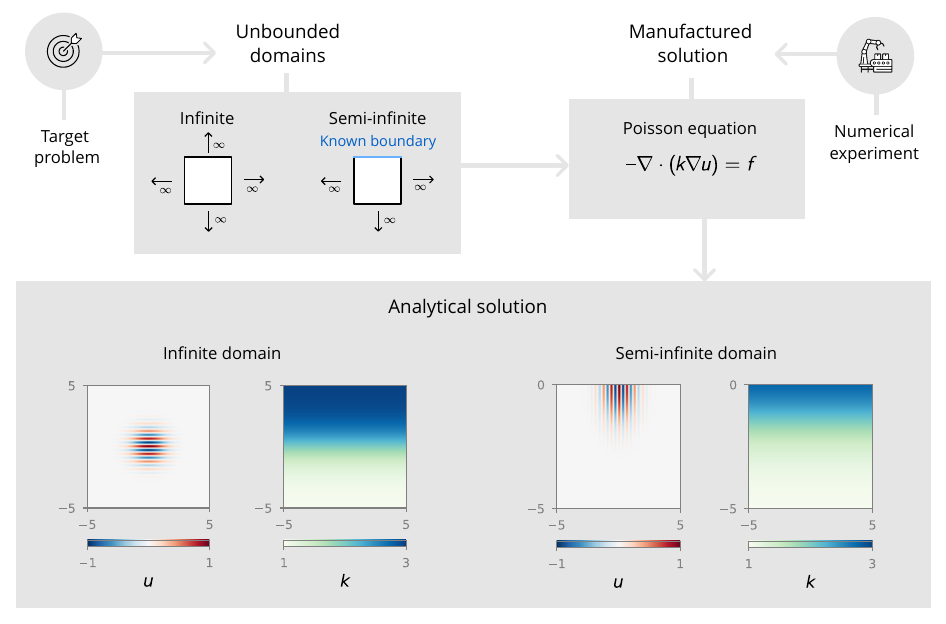}
\caption{Manufactured problem design for unbounded inverse problems. (Top) Schematic workflow illustrating the formulation process, progressing from target definition and domain specification (infinite and semi-infinite) to the generation of a manufactured solution satisfying the Poisson equation $-\nabla \cdot (k \nabla u) = f$. (Bottom) Analytical solutions $u$ and spatially varying coefficient fields $k$. The infinite-domain configuration (left) utilizes parameters $\alpha=0.5$, $\beta=5$, and $\epsilon=1$. The semi-infinite-domain configuration (right) utilizes $\alpha=0.5$, $\beta=5$, and $\epsilon=0.75$.}
    \label{fig:analytical_solution}
\end{figure*}

To evaluate our proposed methodology, two benchmark inverse problems are considered: one defined on an infinite domain and another on a semi-infinite domain. These problems are designed to reproduce challenges associated with applying PINNs to unbounded regions: the solution approximation with absence of explicit boundary information, the recovery of unknown spatially varying parameters and the representation of the solution behavior outside the region of the training. Both problems are formulated as inverse problems based on the Poisson equation,
\begin{equation*}
    -\nabla \cdot (k(x,y)\nabla u(x,y)) = f(x,y),
\end{equation*}
where $u(x,y)$ denotes the unknown solution field, $k(x,y)$ represents the
spatially varying coefficient to be recovered, and $f(x,y)$ is the source
term. The objective is to simultaneously approximate the solution field $u$
and reconstruct the unknown coefficient field $k(x,y)$ using the governing
equation together with sparse observations of both fields. To generate
controlled test cases with known solutions, the method of manufactured
solutions is employed \citep{shunn2007manufactured}. This approach allows the
analytical solution and the coefficient field to be prescribed, while the
source term $f(x,y)$ is constructed such that the governing equation is
exactly satisfied.\\

\textbf{Infinite domain:} the analytical solution and the spatially varying
coefficient are defined as
\begin{equation*}
    u(x,y) = e^{-\alpha (x^2+y^2)} \cos(\beta y), 
\end{equation*}
and 
\begin{equation*}
k(y) = 1 + \frac{2}{1+e^{-y/\varepsilon}}.
\end{equation*}

The chosen solution presents two characteristics for the proposed
problem: it contains oscillatory behavior near the region of interest and
gradually decays as the distance from the origin increases. Similarly, the
coefficient $k$ approaches constant values as $y \rightarrow \pm \infty$,
representing a uniform physical property in the far field. These properties allow the evaluation of whether the neural network can capture the behavior of the solution in regions where training information becomes increasingly sparse.\\

\textbf{Semi-infinite domain:} the same approach is extended by introducing a Dirichlet boundary condition on the finite boundary of the domain, while the remaining directions remain unbounded. This configuration represents a more challenging scenario because the network must satisfy explicit boundary information while simultaneously learning the asymptotic behavior of the solution. The analytical solution is defined as

\begin{equation*}
\begin{cases}
    u(x,y) = e^{-\alpha (x^2+y^2)} \cos(\beta x), & (x,y) \in \Omega, \\[6pt]
    u(x,0) = e^{-\alpha x^2} \cos(\beta x), & x \in \partial \Omega_y ,
\end{cases}
\end{equation*}
with
\begin{equation*}
    k(y) = 1 + \frac{2}{1+e^{-(y+1.5)/\varepsilon}}.
\end{equation*}

The semi-infinite formulation maintains similar solution characteristics to the
infinite case, allowing a direct comparison while introducing the additional
difficulty associated with boundary constraints in partially unbounded domains.
The analytical solutions and the corresponding spatially varying coefficient
fields for both benchmark problems are illustrated in
Fig.~\ref{fig:analytical_solution}. The figure highlights the spatial decay
of the manufactured solutions, the oscillatory behavior near the region of
interest, and the asymptotic behavior of the coefficient field in the far
field. These characteristics motivate the proposed sampling strategies for
unbounded domains. Although these benchmark problems enable controlled and
quantitative evaluation of the proposed methodology, they represent a
simplified setting compared with the complexity of practical PDE-based
applications. A current challenge in physics-informed neural networks is the
scaling of these methods from controlled benchmark problems to more complex
and application-oriented settings involving higher-dimensional domains,
complex geometries, heterogeneous media, and observational data with noise. The present
formulation is therefore intended primarily as a controlled test for
examining the proposed sampling strategies and network architectures, where
the effects of individual methodological choices can be isolated and
quantified. Further studies involving more complex governing equations,
realistic geometries, and field observations are required to assess the
transferability of these findings to practical applications.

\section{KAN hyperparameters optimization}\label{sec:hyperparameters}

The KAN hyperparameters were optimized using the Optuna framework within a bilevel optimization strategy \cite{akiba_optuna_2019}. In this strategy, the upper level evaluates candidate hyperparameter sets based on the mean error. The lower level trains the network with those hyperparameters to optimize the network parameters. The mean error was computed as the arithmetic mean of the relative $L_2$ errors of the solution field $u$ and the coefficient field $k$, evaluated over the domain $[-10,10]^2$.

\begin{figure*}[ht]
    \centering
    \includegraphics[width=6.4 in]{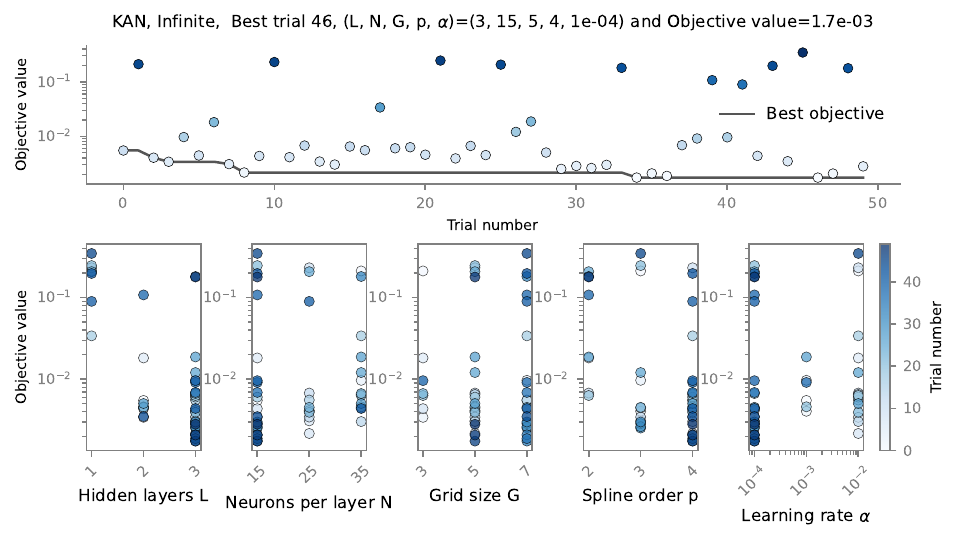}
\caption{Hyperparameter optimization results for the PIKAN model. The top panel shows the mean global error obtained in each trial, with the solid line indicating the best objective value achieved up to each trial. The lower panels show the corresponding error as a function of the number of hidden layers L, neurons per layer N, grid size G, polynomial order p, and learning rate $\alpha$. The color of each marker indicates the trial number, illustrating the progression of the optimization toward lower-error configurations. All error values are shown on a logarithmic scale.}
    \label{fig:kan_infinite_hyperparameter_optimization}
\end{figure*}

The search considered five hyperparameters: the number of hidden layers,
neurons per hidden layer, grid size, spline order, and learning rate. For the
MLP architectures, the activation function was also included as a
hyperparameter. The optimization process and the influence of these
hyperparameters on the objective value are illustrated in
Fig.~\ref{fig:kan_infinite_hyperparameter_optimization}. We identified configurations for
the infinite and semi-infinite problems, shown in
Table~\ref{tab:optimal_hyperparameters}.

The configurations identified did not use the maximum values of layers, neurons, or grid size. This indicates that the search ranges were sufficient to find configurations without requiring the largest model sizes. Additional optimization results are provided in Supplementary Material (Section~S1, Complementary hyperparameter optimizations). 

In addition to the architectural and optimization hyperparameters, the relative
weighting of the loss terms was considered because the inverse formulation
requires the simultaneous optimization of the solution field $u$ and the
unknown coefficient field $k$. Fixed loss weights can lead to an imbalance
between these objectives because their loss components can evolve at
different rates during training. In particular, continued minimization of one
loss component may lead to overfitting of the corresponding constraint while
providing limited improvement, or even degradation, in the reconstruction of
the other field. Adaptive weighting was therefore used to dynamically adjust
the relative contributions of the loss terms during optimization. This
provides a mechanism to compensate for changes in the relative optimization
progress of $u$ and $k$ and to reduce the dominance of individual loss
components during training.

The comparison between fixed and adaptive weighting strategies is reported
in the Supplementary Material (Section~S2, Adaptive weighting). The results
show that the adaptive strategy maintains a solution error comparable to that
obtained with the prescribed weighting schedule while reducing the error in
the reconstructed coefficient $k$. These results indicate that adaptive
weighting can mitigate the imbalance between the competing objectives and
help prevent excessive optimization of one component at the expense of the
other. Based on these results, adaptive weighting was used in the inverse
experiments reported below.

\begin{table*}[ht!]
    \centering
    \sffamily\small
\caption{Optimal hyperparameter configurations.
Here, L denotes the number of hidden layers, N the number of neurons per
layer, G the KAN grid size, $\sigma$ the activation function, p the
polynomial order, and $\alpha$ the learning rate. The objective value
corresponds to the mean global error.}
    \label{tab:optimal_hyperparameters}
    \begin{tabular}{llcccccc}
        \hline
        Model & Domain & L & N & G/$\sigma$ & p &
        $\alpha$ & Objective value \\
        \hline
        KAN & Infinite       & 3 & 15  & 5       & 4 & $1.00\times10^{-4}$ & $1.75\times10^{-3}$ \\
        KAN & Semi-infinite & 3 & 15  & 5       & 3 & $1.00\times10^{-2}$ & $8.38\times10^{-4}$ \\
        MLP & Infinite       & 3 & 90  & Sine    & -- & $1.00\times10^{-3}$ & $1.32\times10^{-3}$ \\
        MLP & Semi-infinite & 2 & 104 & Sigmoid & -- & $1.00\times10^{-3}$ & $1.97\times10^{-3}$ \\
        \hline
    \end{tabular}
\end{table*}

\section{Sampling strategies}\label{sec:sampling}

The proposed methodology incorporates prior physical knowledge through the distribution of the training points. Since the objective is to solve inverse problems in unbounded domains, where the solution typically exhibits localized features and asymptotic decay, the sampling strategy determines which regions contribute most strongly to the optimization process. Previous studies have shown that the distribution of collocation points has a substantial impact on the convergence and accuracy of physics-informed neural networks, often determining whether the optimization process successfully captures the underlying solution behavior \citep{wu_comprehensive_2023}. Consequently, evaluating the influence of the sampling distribution is relevant to determine whether introducing this inductive bias improves the approximation of both the solution field and the unknown coefficient. In contrast to residual-based adaptive sampling strategies, which iteratively redistribute collocation points according to the PDE residual during training \citep{wu_comprehensive_2023}, the proposed approach employs a fixed physics-informed sampling distribution derived from prior knowledge of the expected asymptotic behavior of the solution. This strategy embeds physical insight into the training process from the outset, without requiring adaptive updates throughout the optimization.

\begin{figure*}[ht]
    \centering
    \includegraphics[width=6.3 in]{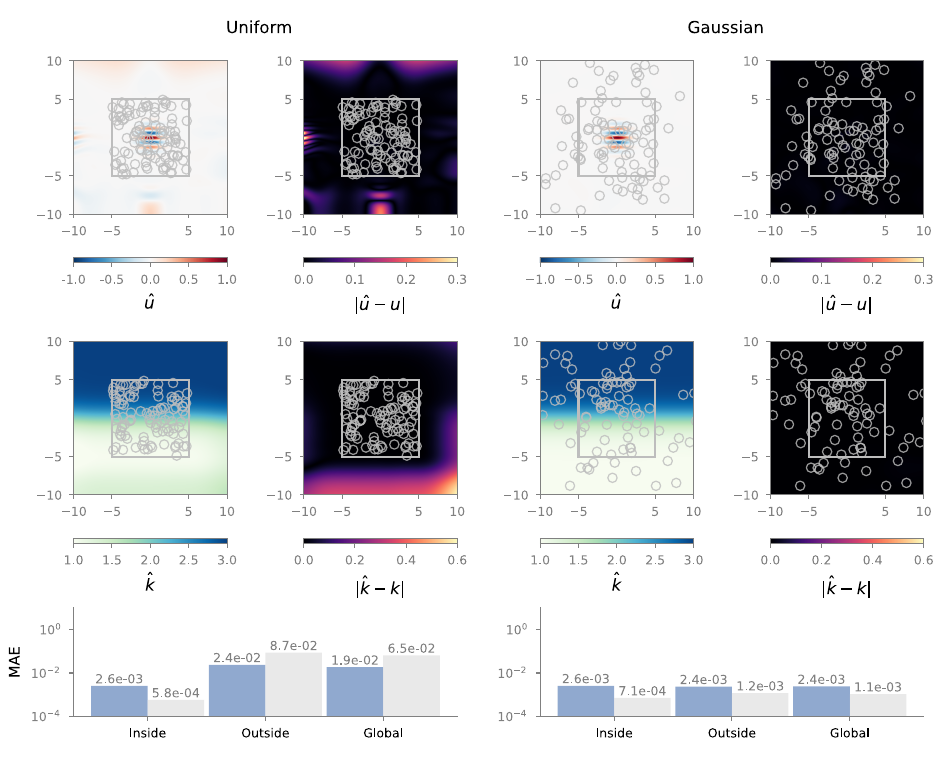}
\caption{Spatial generalization of the proposed approach for the infinite-domain inverse problem under uniform and Gaussian sampling. The predicted fields and absolute errors for $u$ and $k$ are shown together with the MAE errors measured inside, outside, and over the complete evaluation domain. The gray rectangle marks the training region $[-5,5]\times[-5,5]$ within the evaluation domain $[-10,10]\times[-10,10]$. Gray circles indicate the observational points, which are sampled from the same distribution as the PDE collocation points.}
    \label{fig:infinite_sampling}
\end{figure*}

\begin{figure*}[ht]
    \centering
    \includegraphics[width=6.3 in]{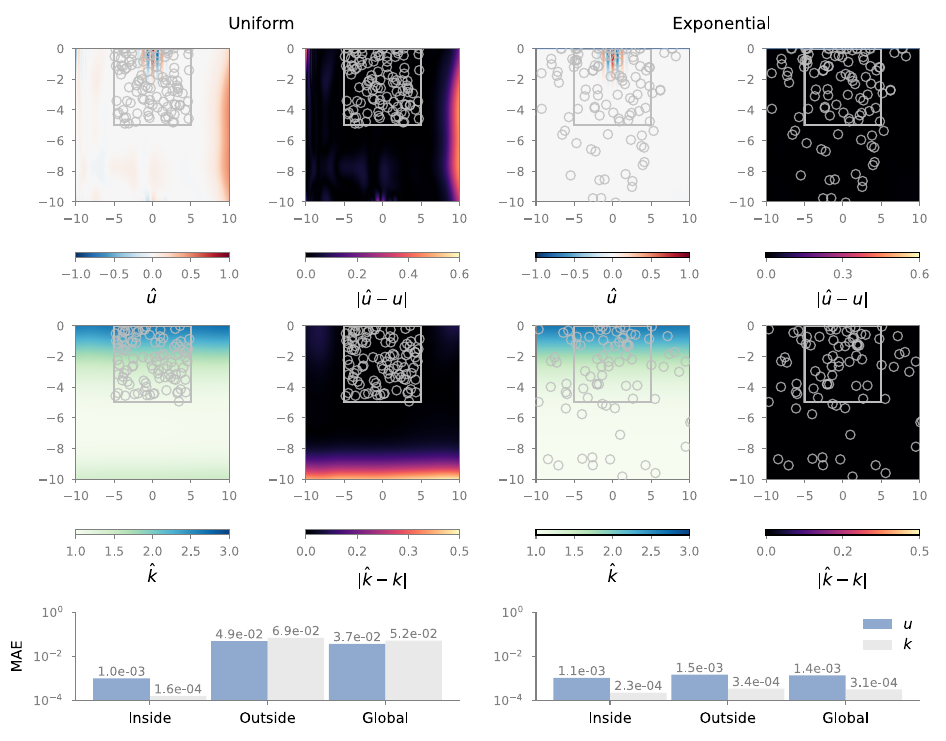}
    \caption{Sampling-strategy comparison for the semi-infinite-domain inverse problem. Uniform and exponential sampling are evaluated through the predicted fields, absolute errors, and MAE inside, outside, and across the evaluation domain. The training region is $[-5,5]\times[-5,0]$ within $[-10,10]\times[-10,0]$, with the top-boundary samples highlighted in blue.}
    \label{fig:semi_infinite_sampling}
\end{figure*}

The effect of the spatial distribution of the training points is illustrated in Figs.~\ref{fig:infinite_sampling} and \ref{fig:semi_infinite_sampling}. The figures compare the spatial predictions and errors obtained using uniform, Gaussian and exponential sampling for the two proposed problems. In particular, the results highlight how concentrating collocation points in the region where the solution has significant spatial variation can achieve similar accuracy in the approximation within the training domain and improve the generalization to the unbounded region outside the training domain.

To ensure a fair comparison, all experiments employ the same neural
architecture, optimizer, loss weights, and total number of training points.
Only the probability distributions used to generate the PDE collocation points
and observational points are modified. Thus, both types of training points are
sampled according to the strategy under evaluation. The different sampling
strategies are evaluated using the MAE of both the reconstructed solution $u$
and the recovered coefficient $k$, together with the convergence behavior of
the loss function during training. This analysis allows us to assess whether
incorporating prior knowledge through non-uniform sampling provides a
measurable advantage over conventional uniform sampling for inverse problems
defined on unbounded domains.

The proposed sampling strategy assumes that the solution exhibits localized behavior together with monotonic or exponential decay toward infinity. Problems characterized by persistent oscillations, algebraic decay, or multiple separated regions of activity may require alternative probability distributions or adaptive sampling strategies. Therefore, the present approach should be viewed as a physics-guided sampling strategy tailored to problems whose asymptotic behavior is known a priori rather than as a universal sampling methodology. 

The present study assumes the availability of observational data not only on the domain boundaries but also at interior locations. This assumption is consistent with applications in which interior data are obtained through direct exploration \cite{lutenegger_situ_2021}. However, interior observations in practical inverse problems are typically sparse and located at discrete points determined by cost and accessibility, rather than sampled from a prescribed distribution as in the present experiments. In other applications, such as seismic imaging, measurements are acquired mainly at receivers located on or near the Earth's surface, and interior observations are not available \citep{tarantolaInverse}. Consequently, extending the proposed framework to scenarios with fewer interior observations and with irregular observation locations is a direction for future research and would facilitate its transfer to real-world large-scale inverse problems involving naturally unbounded domains. Future work will investigate whether the proposed sampling strategy can compensate for reduced interior observations through stronger physics constraints or adaptive collocation strategies.

\section{PIKANs trade-offs}\label{sec:pikans}

Although Kolmogorov--Arnold Networks have recently attracted attention as an alternative to multilayer perceptrons, their practical applicability to physics-informed inverse problems depends not only on solution accuracy but also on computational efficiency. Unlike conventional
MLPs, where nonlinear transformations are applied through fixed activation functions at the neurons, PIKANs employ learnable spline functions on the network edges. This formulation introduces additional operations during forward evaluation and automatic differentiation, which can become relevant when higher-order derivatives are required to evaluate PDE residuals
\citep{liu2024kan, shukla_comprehensive_2024}. Therefore, the objective of this analysis is to characterize the trade-off between predictive accuracy, model size, and computational cost for inverse problems in unbounded domains.

\begin{figure*}[htbp]
\centering
\includegraphics[width=6.3 in]{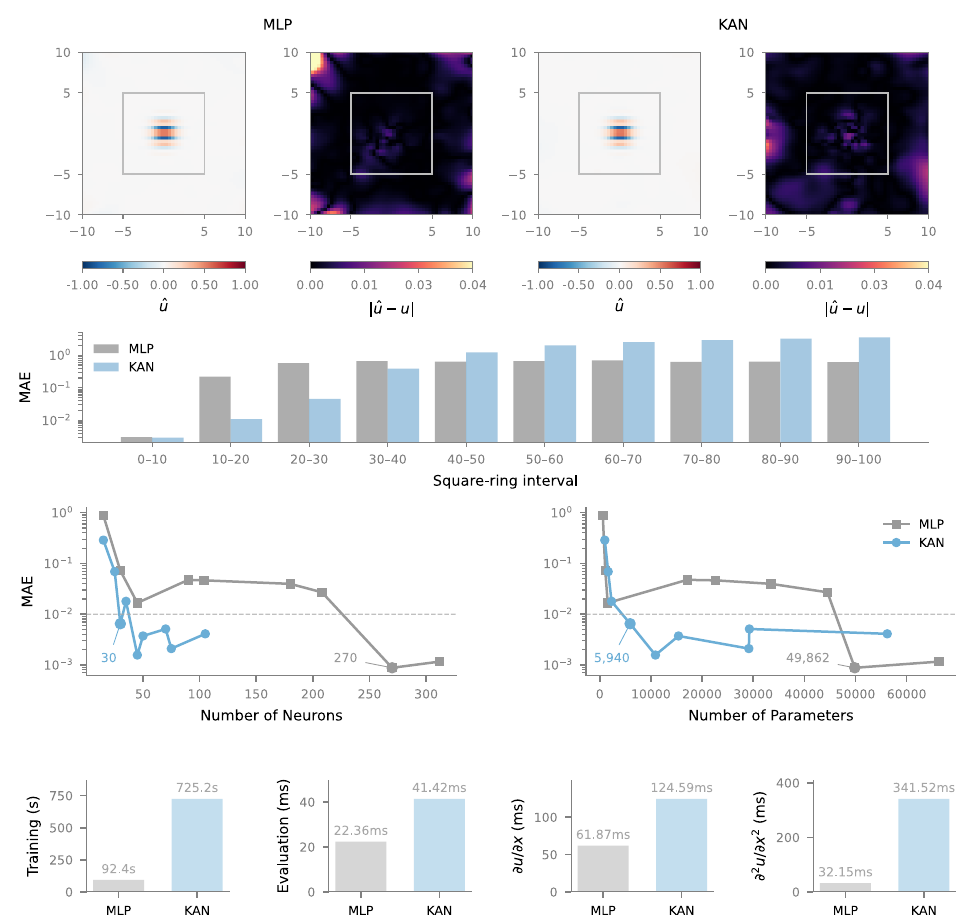}
\caption{Accuracy and computational-cost comparison between the physics-informed MLP and PIKAN approaches for the inverse problem in an unbounded domain. The top row shows the predicted solution $\hat{u}$ and corresponding absolute error $|\hat{u}-u|$ for both approaches in the far field, evaluated over $[-10,10]^2$, with the training domain indicated by the gray square. The second row shows the MAE across square-ring intervals extending from the training region to $[-100,100]^2$. The third row presents the MAE as a function of the number of hidden neurons (left) and trainable parameters (right). The bottom row compares the computational costs in terms of training time, forward evaluation time, and the time required to compute first- and second-order derivatives.}
\label{fig:accuracy_efficiency_tradeoff}
\end{figure*}

To quantify this trade-off, the optimized MLP and PIKAN configurations are
compared under the same training conditions. The analysis considers solution
accuracy, number of trainable parameters, training time, inference time, and
the cost of computing first- and second-order derivatives. These quantities
provide complementary measures of the computational requirements of each
architecture. In particular, comparing accuracy with both neuron count and
parameter count allows us to determine whether the reduced number of neurons
typically required by KANs translates into a smaller model or improved
computational efficiency.

Fig.~\ref{fig:accuracy_efficiency_tradeoff} shows that both architectures
achieve similar global error levels over comparable ranges of trainable
parameters. However, the relationship between accuracy and network size
differs between the two architectures. PIKANs achieve low error with fewer
hidden neurons, which is consistent with the additional flexibility provided
by their learnable edge functions. This reduction in neuron count does not
necessarily translate into fewer trainable parameters or lower computational
cost, since the spline-based edge functions introduce additional learnable
parameters and require more operations during forward and derivative
evaluation.

The training-time comparison shows that the PIKAN requires more time to
reach its final solution than the MLP under the same training procedure.
Similarly, forward evaluation and derivative computation are more
computationally demanding for the PIKAN. The difference becomes more
pronounced for second-order derivatives, which are required for the PDE
residual in the present problem. This behavior is consistent with the
additional computational operations introduced by the spline-based edge
functions and their associated automatic differentiation.

\section{Conclusions}\label{sec:conclusions}

The sampling experiments show that Gaussian and exponential distributions
improve the reconstruction of both the solution and the unknown coefficient,
particularly in the far field, relative to uniform sampling. Unlike uniform
sampling, which requires defining a finite spatial domain to generate
collocation points, the proposed distributions can allocate points according
to their spatial probability density without explicitly restricting the
sampling to a finite computational domain. Because the spatial distribution of
collocation points is governed by explicit distribution parameters, this
strategy provides a direct means of incorporating prior knowledge about the
expected spatial behavior of the solution into the training process. These
parameters can also be optimized for a given problem to adapt the sampling
distribution to the spatial characteristics of the solution. 
 
The comparison between MLPs and PIKANs shows a trade-off between accuracy and computational cost. PIKANs match or exceed the accuracy of MLPs while using fewer neurons, but their spline-based functional representation increases the cost of training, inference, and derivative evaluation. This cost is offset by a structural advantage: whereas MLPs rely on fixed activation functions and trainable weighted connections, KANs place learnable univariate functions on the edges themselves, a design grounded in the Kolmogorov--Arnold representation theorem. Reducing the computational overhead of spline evaluation and differentiation, through more efficient implementations, would make PIKANs more appropriate for large-scale, physics-informed problems.

Future work should aim to scale the methodology to large-scale inverse problems posed on unbounded domains, e.g., seismic velocity inversion and geotechnical site characterization. Further research will also investigate hybrid BEM--PINN formulations, in which boundary element methods enforce radiation conditions at infinity, removing the need to truncate the computational domain artificially. Architectures tailored to the problem and based on orthogonal polynomial bases represent a further promising direction, as they allow known solution structure to be encoded directly into the network. The choice of basis should be guided by the nature of the domain: for unbounded problems, families whose orthogonality weight decays at infinity are generally preferable, since this decay reflects the far-field behavior of the physical solution. Hermite functions, for instance, are well suited to solutions that decay from a localized source, Laguerre functions to radial or half-line problems, and mapped Chebyshev functions to solutions with algebraic decay, as commonly required by radiation conditions in scattering problems.
 
\section*{CRediT authorship contribution statement}

\textbf{Gregorio Pérez-Bernal}: Writing -- original draft, Visualization, Software, Methodology, Investigation, Formal analysis, Data curation, Conceptualization. \textbf{Óscar Rincón Cardeño}: Writing -- original draft, Visualization, Software, Methodology, Investigation, Formal analysis, Data curation, Conceptualization. \textbf{Silvana Montoya-Noguera}: Writing -- review \& editing, Supervision, Funding acquisition, Conceptualization. \textbf{Nicolás Guarín-Zapata}: Writing -- review \& editing, Visualization, Supervision, Software, Project administration, Methodology, Funding acquisition, Formal analysis, Conceptualization.

\section*{Code and data availability}

The code and data used in this study are available in the project repository at \url{https://github.com/oscar-rincon/PIKAN-unbounded-domains} \cite{rincon_cardeno_2026_23087248}.

\section*{Declaration of competing interest}

The authors declare that they have no known competing financial interests or personal relationships that could have appeared to influence the work reported in this paper.

\section*{Acknowledgment}

This work was supported by Universidad EAFIT, Colombia, Program No. 12330032023.

\bibliographystyle{model6-num-names}
\bibliography{refs}

\end{document}